# NeuroMorse: A Temporally Structured Dataset For Neuromorphic Computing

**Ben Walters[†], Yeshwanth Bethi[⋆], Taylor Kergan[‡], Binh Nguyen[♯], Amirali Amirsoleimani[◇], Jason K. Eshraghian[‡], Saeed Afshar[⋆], Mostafa Rahimi Azghadi[†]**

[†] College of Science and Engineering, James Cook University, Townsville, Australia
[⋆] International Centre for Neuromorphic Systems, Western Sydney University, Australia
[◇] Department of Electrical Engineering and Computer Science, York University, Toronto, Canada
[‡] Department of Electrical and Computer Engineering, University of California, Santa Cruz, The United States of America

**Abstract.**
  Neuromorphic engineering aims to advance computing by mimicking the brain's efficient processing, where data is encoded as asynchronous temporal events. This eliminates the need for a synchronisation clock and minimises power consumption when no data is present. However, many benchmarks for neuromorphic algorithms primarily focus on spatial features, neglecting the temporal dynamics that are inherent to most sequence-based tasks. This gap may lead to evaluations that fail to fully capture the unique strengths and characteristics of neuromorphic systems. In this paper, we present NeuroMorse, a temporally structured dataset designed for benchmarking neuromorphic learning systems. NeuroMorse converts the top 50 words in the English language into temporal Morse code spike sequences. Despite using only two input spike channels for Morse dots and dashes, complex information is encoded through temporal patterns in the data. The proposed benchmark contains feature hierarchy at multiple temporal scales that test the capacity of neuromorphic algorithms to decompose input patterns into spatial and temporal hierarchies. We demonstrate that our training set is challenging to categorise using a linear classifier and that identifying keywords in the test set is difficult using conventional methods. The NeuroMorse dataset is available at 10.5281/zenodo.12702379, with our accompanying code at https://github.com/jc427648/NeuroMorse.

## 1. Introduction

Neuromorphic computing draws upon brain-inspired principles in order to develop low power computing paradigms. Compared to current Artificial Neural Networks (ANNs), neuromorphic Spiking Neural Networks (SNNs) rely on the dynamic generation and transmission of binary spikes between neuronal nodes. This is significantly different from the data-driven approaches that are in use today (1; 2). However, this shift from data-driven to event-driven computation requires new methods to train and test neuromorphic models.

Spiking neural networks encode information across both spatial and temporal domains using precisely-timed binary-valued signals, making them well-suited for event-based tasks. However, benchmarking these networks effectively requires datasets that contain this rich spatio-temporal structure. Many current benchmarks fail to leverage temporal data aspects adequately (3). Instead, they often rely on static datasets converted into spatio-temporal spike trains or event-based datasets that lack a focus on the hierarchical structure of spike/event times. Developing benchmarks that provide a comprehensive understanding of all model elements is crucial for evaluating and improving SNNs (4; 5; 6; 7; 8).

Table 1 provides a comparison of various datasets used to benchmark neuromorphic systems. In this Table, we provide a description of each dataset and a qualitative guide to the spatial or temporal hierarchy information contained in each dataset. Here, "High" implies that the data is almost explicitly encoded in this domain, and virtually impossible to decipher by using other means. "Medium" implies that structures and dependencies in this domain do exist, but temporal and spatial features carry similar importance. Event-based gesture datasets are an example of this, where the spatial information outweighs the importance of the temporal information, given that such tasks can be solved without recurrent models. Audio classification, as in Spiking Heidelberg Digits (SHD)/Spiking Speech Commands (SSC), have stronger dependence on time-varying features, though the number of channels is often similar to or greater than that of the number of sequence steps when training classification models. "Low" implies that some of the information is encoded in this domain, however, the majority of the information is encoded in another. An example of this is the N-MNIST (15) dataset that simply converts the MNIST dataset into a neuromorphic dataset via an event based camera. Finally, "None" implies that no data is encoded in this domain.

As shown in Table 1, some common datasets used to investigate neuromorphic architectures are static image datasets, such as MNIST (9; 10), Fasion-MNIST (11), Extended-MNIST (12), CIFAR (13) and Caltech (14). However, these datasets require a transformation from static images to temporal spike sequences. Oftentimes, this is performed by either rate or latency encoding, neither of which emphasises a hierarchy of events in time. Instead, these datasets prioritise learning the spatial hierarchies encoded within the data. In response to these static datasets, some research has been dedicated to developing spatio-temporal datasets that are more suited for neuromorphic computing. This includes an event-based version of MNIST (15), Dynamic Vision Sensor (DVS) gestures (16; 20; 17) and spiking audio datasets like the SHD dataset (19). However, the timing information, even in these event-based datasets, is not crucial to the success of the downstream task. In fact, events can be binned at the input channel with minimal effect on the final result. Networks trained using these datasets often show minimal drop in performance even when the temporal information is completely removed from the spiking datasets as shown by (3).

Table 1: Comparison of various datasets used to verify neuromorphic computing architectures

| Dataset | Description | Spatial Hierarchies | Temporal Hierarchies |
|---|---|---|---|
| **Static Datasets** | | | |
| MNIST (9; 10) | Grayscale images of handwritten digits (0-9) | High | None |
| Fashion-MNIST (11) | Grayscale images of clothing items (10 categories) | High | None |
| E-MNIST (12) | Grayscale images of handwritten alphanumeric characters | High | None |
| CIFAR (13) | Colour images of various objects | High | None |
| Caltech (14) | Colour images of various objects | High | None |
| **Event-Based Datasets** | | | |
| N-MNIST (15) | Spiking version of MNIST dataset | High | Low |
| Poker DVS (16) | Poker pip symbols recorded through DVS cameras | High | Low |
| DVS Gestures (17) | Hand gestures recorded by DVS cameras | High | Medium |
| ASL DVS (18) | American Sign Language letters recorded via DVS cameras | High | Medium |
| SSC (19) | Spiking version of the speech commands dataset | Medium | Medium |
| SHD (19) | Spike-based audio classification of spoken digits | Medium | Medium |
| **NeuroMorse** | Top 50 English words converted to Morse code spike sequences | Low | High |

In this paper, we present NeuroMorse, a dataset where the temporal aspects of spike-based learning are more critical than the spatial aspects. We reduce the number of input channels, i.e. the spatial aspect of the data, to just two, whilst encoding the data in the form of spike sequences. To the best of our knowledge, NeuroMorse is the only neuromorphic dataset that prioritises temporal data representation over spatial or channel-wise representations. This focus could play a crucial role in advancing the development of more effective neuromorphic learning algorithms and architectures.

The structure of the paper is as follows: Section 2 describes the generation and formatting of the NeuroMorse dataset and details its specific nature. Section 3 discusses our preliminary analysis of the dataset, while Section 4 provides a summary of the paper

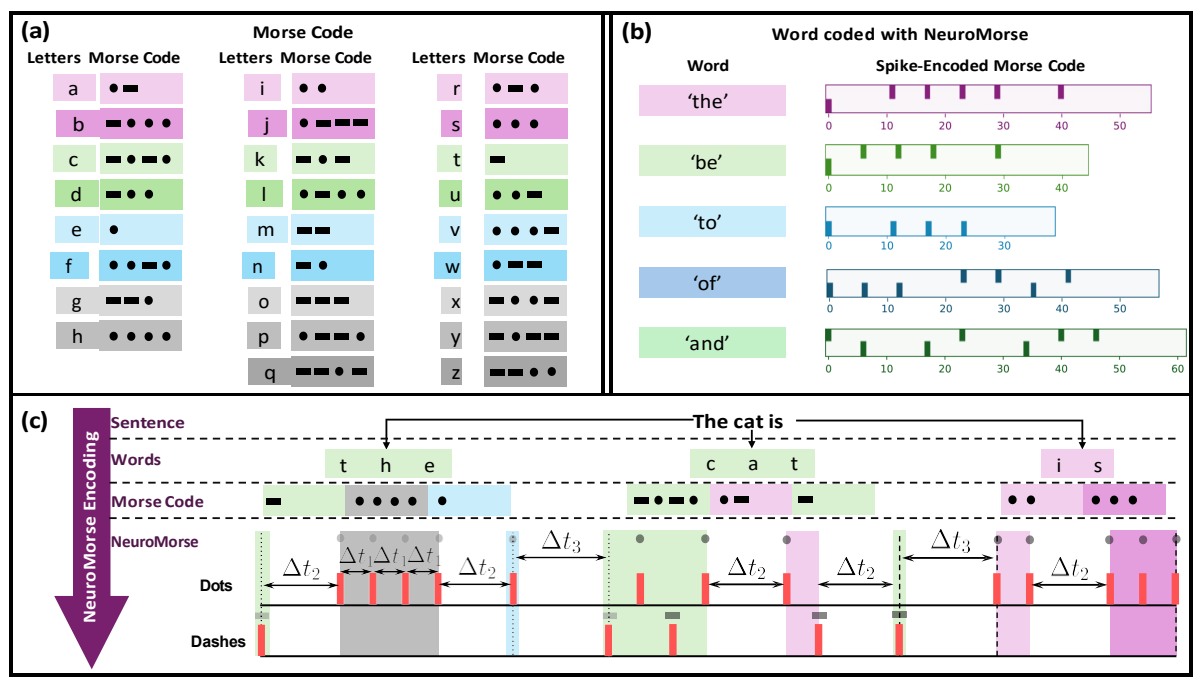

Figure 1: Conversion of Morse code to spiking sequence (a) An example of the current standard of Morse code. (b) Example spikes sequences of the top 5 words in the English language. Note the 15 blank timesteps at the end of each sequence to distinguish each word. (c) An example of the conversion process from a sentence to a spike sequence. Here $\Delta t_1$ represents the timesteps between consecutive Morse characters (dots and dashes) and was set to 5, $\Delta t_2$ represents the timesteps between consecutive english characters and $\Delta t_3$ represents the timesteps between consecutive English words.

and our findings.

## 2. Methodology

To begin creating our NeuroMorse dataset, we took inspiration from a Morse code dataset (21). Morse code has two input representations: dots and dashes, where the sequence of dots and dashes represents alphanumeric characters, as shown in Figure 1 (a). Thus, these sequences are easily transferable to the temporal domain, where two input channels represent dots and dashes. Figure 1(b) illustrates this by displaying the top five most frequently used English words translated into NeuroMorse. Figure 1(c) presents more details at the sentence and word encoding levels. Here, $\Delta t_1$, $\Delta t_2$ and $\Delta t_3$ represent the timestep interval between consecutive dots and dashes (5 timesteps), between consecutive characters (10 timesteps) and between consecutive words (15 timesteps) respectively.

The NeuroMorse training set includes the top 50 most frequently used words in the English language, which have been transformed into spike sequences. Each entry in the dataset provides the channel, spike time, and labels corresponding to each word. For the test set, spike sequences were generated from a corpus of 50,441 Simple Wikipedia articles (with punctuation removed for convenience). The primary metric for this dataset is identifying occurrences of the top 50 words (referred to as keywords) in the test set. Additionally, to aid in keyword identification, the dataset includes start and end times for each keyword, along with the spike time, channel, and label data.

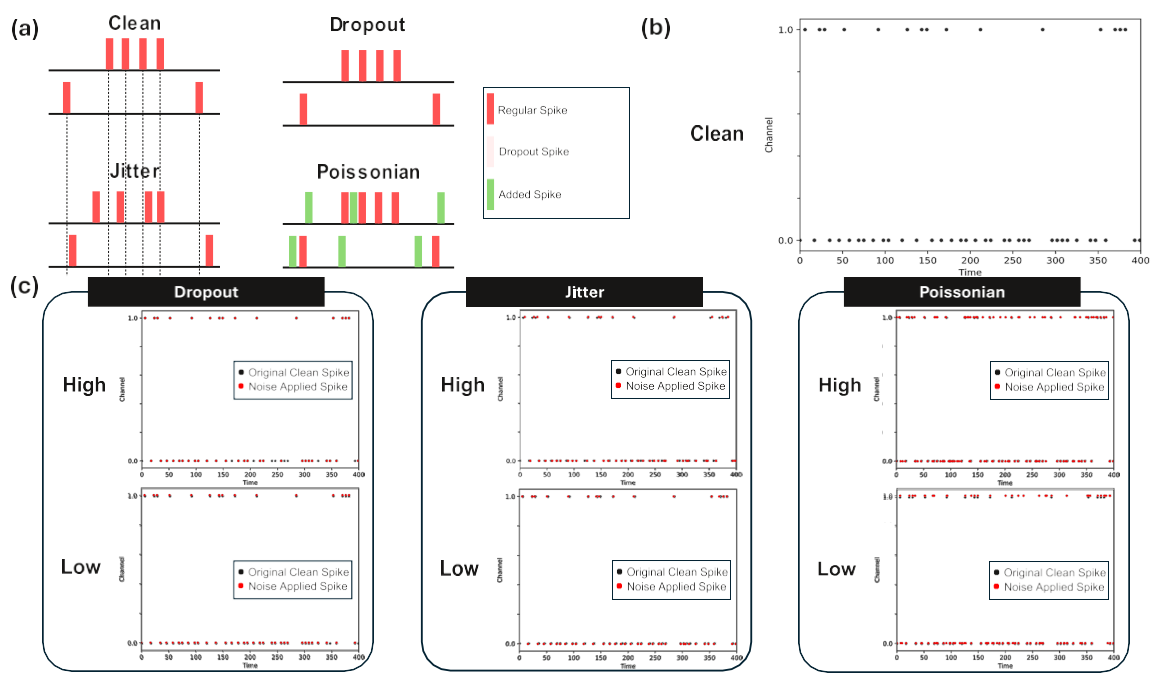

Figure 2: (a) Various types of noise are introduced into the dataset. (b) A sample spike sequence from the test set. (c) Comparison of noisy and non-noisy dataset samples of the test set in (b).

To increase the challenge of the dataset, we introduced noisy versions of both the training and test sets. Three types of noise were applied, as illustrated in Figure 2(a). The first type, spike removal (dropout), involved randomly omitting spikes with fixed probabilities of 3.33% and 6.67% for the low and high noise cases, respectively. The second type, jitter, was introduced by adding noise sampled from a Gaussian distribution (with standard deviations of 1 and 2 timesteps for low and high cases) to each spike. The final type involved adding Poissonian spike trains with rates of 0.05 and 0.1 per timestep for low and high noise cases. Thus, the three possible cases of noise level (None, Low and High) for each of the three types of noise results in 27 total datasets. Examples of clean and noisy data can be seen in Figure 2 (b) and (c). Furthermore, we provide examples of keywords being identified in a sentence in the test set in Figure 3(a) and (b).

The training and testing datasets are provided in Hierarchical Data Format 5 (HDF5) format for efficient storage and access. Figure 3(c)(i) illustrates the hierarchical structure of the training set. The 'Spikes' category contains the time stamps and channel information for each spike, while the labels are stored in a separate group. The test set follows a similar structure as shown in Figure 3(c)(ii). Additionally, to facilitate evaluation, the labels group in the test set includes the start and end times of each

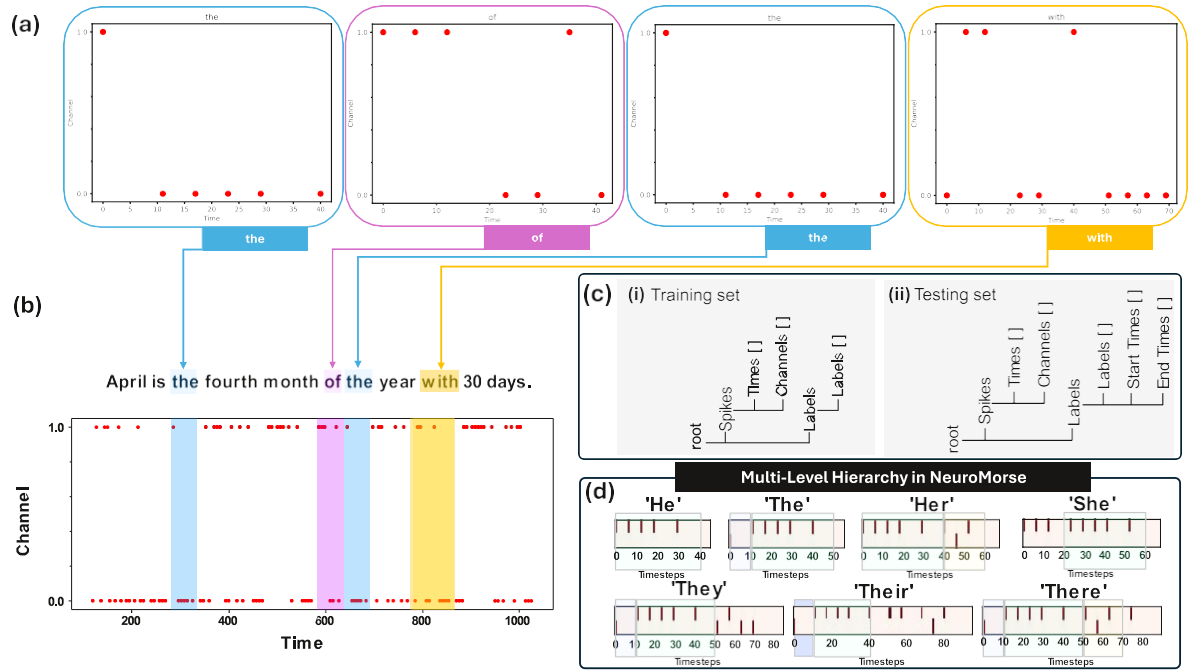

Figure 3: (a) Keyword example spike sequences. (b) Spike sequence of a sentence in the test set containing three different keywords. (c) Summary of the directory trees for (i) the training and (ii) the testing sets in HDF5 format. (d) Examples of the multi-layer hierarchies that exist within some training set examples.

keyword.

NeuroMorse is designed to evaluate the capability of neuromorphic computing architectures to recognise temporal patterns and hierarchical structures within spike sequences. Figure 3 (d) presents sample spike sequences from the training set, highlighting embedded patterns across different words. The overlap of these patterns is particularly pronounced in the test set, which features a more extensive vocabulary. This complexity, arising from hierarchical structures, poses a significant challenge for neuromorphic networks, demanding their ability to accurately identify both individual spike sequences and their relationships within larger patterns.

## 3. Dataset Tests

To show the difficulty in spike sequence learning using standard neuromorphic architectures, we ran our training set through a neuromorphic linear classifier. For this, each dash or dot channel spike sequence was connected to one leaky integrate and fire neuron, as shown in Figure 4 (a). Equation 1 shows the formulation of this neuron model.

$$U[t+1] = \beta U[t] + X[t] \tag{1}$$

Here, $\beta$ is a leak term set to 0.95, $U[t]$ represents the membrane potential at time $t$ and $X[t]$ represents the sum of the input spike channels at time $t$ (where a spike is represented by the value of 1 and 0 otherwise). Snntorch (22) was used to simulate the network. For this task, we ignored the threshold of the neuron and only monitored the membrane potential, which was recorded after each of the 50 training input presentations. The membrane potential vector for all input words was then passed through an ordinary least squares regression module, where each input in the training set was one-hot encoded. We achieved a classification accuracy of 2/50 or 4%, i.e. double the random baseline.

Table 2: Number of keyword instances for each label in the test set.

| Number | Label | Count | Number | Label | Count |
|---|---|---|---|---|---|
| 1 | "the" | 398,449 | 26 | "they" | 23,969 |
| 2 | "be" | 20,136 | 27 | "we" | 1,960 |
| 3 | "to" | 99,718 | 28 | "say" | 1,564 |
| 4 | "of" | 190,969 | 29 | "her" | 7,155 |
| 5 | "and" | 138,885 | 30 | "she" | 8,446 |
| 6 | "a" | 145,727 | 31 | "or" | 26,804 |
| 7 | "in" | 166,335 | 32 | "an" | 21,667 |
| 8 | "that" | 42,578 | 33 | "will" | 4,765 |
| 9 | "have" | 15,464 | 34 | "my" | 605 |
| 10 | "I" | 2,368 | 35 | "one" | 14,914 |
| 11 | "it" | 59,406 | 36 | "all" | 8,351 |
| 12 | "for" | 38,823 | 37 | "would" | 4,292 |
| 13 | "not" | 16,982 | 38 | "there" | 12,819 |
| 14 | "on" | 34,320 | 39 | "their" | 11,654 |
| 15 | "with" | 27,835 | 40 | "what" | 3,320 |
| 16 | "he" | 34,707 | 41 | "so" | 5,568 |
| 17 | "as" | 35,499 | 42 | "up" | 5,565 |
| 18 | "you" | 1,968 | 43 | "out" | 4,619 |
| 19 | "do" | 4,476 | 44 | "if" | 4,923 |
| 20 | "at" | 20,342 | 45 | "about" | 10,929 |
| 21 | "this" | 20,556 | 46 | "who" | 11,351 |
| 22 | "but" | 14,010 | 47 | "get" | 2,766 |
| 23 | "his" | 20,094 | 48 | "which" | 15,791 |
| 24 | "by" | 33,694 | 49 | "go" | 2,117 |
| 25 | "from" | 26,787 | 50 | "me" | 421 |
| | | | | **Total** | 1,768,231 |

We also evaluated the difficulty of detecting spike sequences within the test set. Given that the test set has additional null classes, we trained a basic network as shown in Figure 4 (b), where fifty neurons were connected to the 2 input channels in an all-to-all fashion via weight dependant Spike Timing Dependant Plasticity (STDP) synapses. These synapses only consider causal relationships i.e. pre-synaptic events that occur before post-synaptic events for weight update. Equation 2 illustrates the rule used

$$\Delta w = \eta(1 - \frac{\Delta t}{20} - w), \quad (2)$$

where $\Delta w$ represents the change in synaptic weight, $\eta$ is a scaling factor set to 0.1, $\Delta t$ is the time difference between pre- and post-synaptic spikes and $w$ is the current value of synaptic weight.

Lateral inhibition was added to facilitate a winner-takes-all mechanism, allowing the neurons to discern different patterns in the training set. Furthermore, we included a threshold regulation scheme to ensure all neurons exhibit equal activity levels. The equation that governs this threshold regulation scheme is shown in Equation 3,

$$\Delta V_{th} = A_{th} S[t] - \tau_{th} \qquad (3)$$

where $V_{th}$ is the threshold voltage, $A_{th}$ is a positive fixed value of threshold increase, $S[t]$ is one if the neuron spikes and 0 otherwise and $\tau_{th}$ is a fixed positive value for threshold decay. Here we opted for values of $1 \times 10^{-1}$ and $1 \times 10^{-4}$ for $A_{th}$ and $\tau_{th}$, respectively, and did not perform any parameter optimisation.

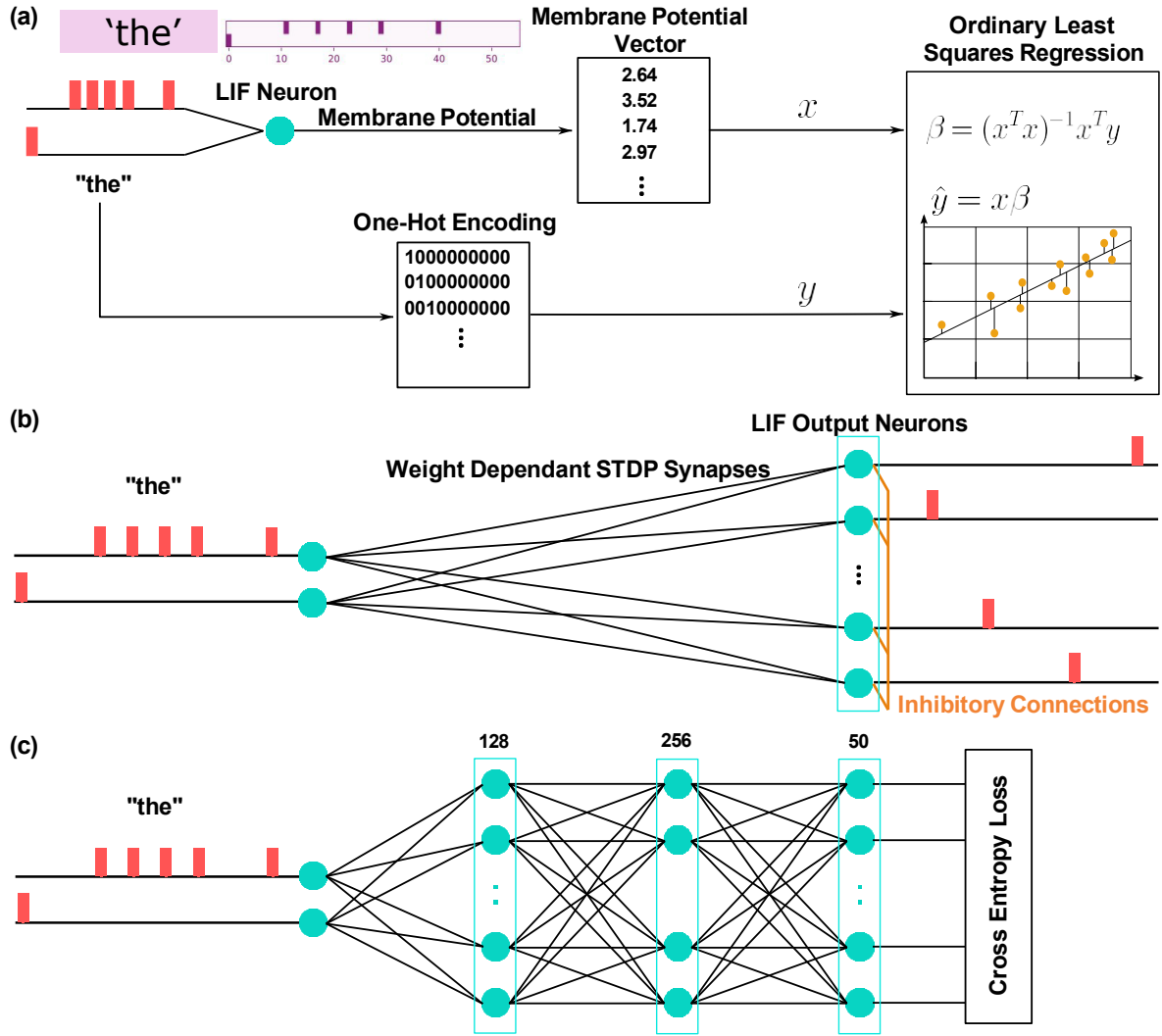

Figure 4: Network structures used to analyse the dataset. (a) A linear classification setup was used to discriminate the training set examples from each other. The 50 words in the training set were presented to the LIF neuron, each generating a membrane potential. Additionally, the words were encoded using a one-hot encoding scheme. (b) A fully connected network to evaluate the test set. (c) A three layer network trained using supervised learning techniques to evaluate both the training and testing sets.

During training, the spike sequences of the 50 training words were presented to the network in a random order. Subsequently, a classification stage was conducted to assign each of the 50 output neurons to a specific class. This involved processing each training sample again (without updating weights or thresholds) and recording the spiking output of each neuron. The output neuron with the highest spike count was assigned the corresponding class. The test set was then introduced to the trained network. For each output neuron, the timing of its spikes was compared to the end times of the correct keywords in the test set. A spike was considered correct if its time matched the end time of the associated keyword. Otherwise, it was classified as incorrect. Our evaluation revealed a total of $97,275$ correct spikes and $50,710,039$ incorrect spikes, indicating a classification accuracy of only 0.19%. Given that the test data contains $1,768,231$ instances of keywords (as shown in Table 2), this highlights a significant challenge: the misclassification of null classes.

Table 3: Summary of dataset tests performed

| Training Method | Training Set Accuracy | Test Set Accuracy | Filtered Test Set Accuracy |
|---|---|---|---|
| Linear Classifier | 4% | N/A | N/A |
| STDP Network | N/A | 0.19% | N/A |
| Supervised Network | 66% | 4.25% | 12.46% |

Additionally, we performed one more test using a more complex SNN to evaluate the complexity of NeuroMorse. Here we utilised spiking backpropagation using surrogate gradients, on a deeper 3-layer network as shown in Figure 4(c). The use of this supervised learning scheme was expected to further improve the accuracy achieved. Furthermore, a deeper network can extract more complex features than the previously utilised shallow networks. To train this network, all inputs were padded to the same length, and the fast sigmoid function with a slope of 15 was used for the surrogate gradients.

After training the network for 2000 epochs, we first presented the training set one more time to identify the keywords. Keywords were identified with 66% accuracy, suggesting that discriminating keywords in the training set is very difficult, even without additional words not present in the training set. The test set was then presented to the network, where keywords were identified with 4.25% accuracy. Finally, we filtered the test set so that non-keywords were removed and only keywords remained (resulting in a random sequence of keywords), and achieved a 12.46% accuracy.

Hence, with three different dataset tests, we show the limited ability to learn temporal hierarchies in current neuromorphic architectures. These results have been summarised in Table 3. We have shown that our training set keywords are difficult to discriminate using a basic linear classifier. We also demonstrate that a basic STDP network is unable to effectively learn and pick out the keywords in the test set. Finally, we have shown that even supervised approaches struggle to discriminate and identify keywords in the training and testing sets. All tests have been performed without noise, which adds to the complexity of the dataset.

## 4. Conclusion

This paper introduced a novel spiking dataset designed to prioritize temporal structures and hierarchies over spatial hierarchies. Our initial experiments demonstrated the dataset's non-linearity, as evidenced by the low accuracy (4%) achieved with ordinary least squares regression. Furthermore, basic neuromorphic networks struggled on the complex NeuroMorse test set, reaching very low accuracies. Given the dataset's emphasis on temporal patterns and hierarchical relationships, it presents a valuable resource for researchers investigating the capabilities of neuromorphic architectures in handling complex temporal data. We encourage the neuromorphic community to utilize this dataset in conjunction with other benchmarks to explore the full potential of their models in tasks such as temporal sequence recognition and hierarchical feature extraction.